\documentclass[conference]{IEEEtran}

\usepackage{amsmath,amsfonts,bm}

\def\eqref#1{equation~\ref{#1}}

\def\1{\bm{1}}

\def\vw{{\bm{w}}}

\def\mA{{\bm{A}}}

\def\mG{{\bm{G}}}
\def\mH{{\bm{H}}}

\def\mW{{\bm{W}}}
\def\mX{{\bm{X}}}

\DeclareMathAlphabet{\mathsfit}{\encodingdefault}{\sfdefault}{m}{sl}
\SetMathAlphabet{\mathsfit}{bold}{\encodingdefault}{\sfdefault}{bx}{n}

\def\gB{{\mathcal{B}}}

\def\gM{{\mathcal{M}}}
\def\gN{{\mathcal{N}}}
\def\gO{{\mathcal{O}}}
\def\gP{{\mathcal{P}}}

\def\gR{{\mathcal{R}}}

\def\gT{{\mathcal{T}}}

\def\gW{{\mathcal{W}}}

\def\sA{{\mathbb{A}}}

\newcommand{\Ls}{\mathcal{L}}
\newcommand{\R}{\mathbb{R}}

\usepackage{hyperref}
\usepackage{url}            
\usepackage{booktabs}       
\usepackage{microtype}      
\usepackage{xcolor}         

\usepackage{microtype}
\usepackage{graphicx}
\usepackage{subfigure}
\usepackage{url}
\usepackage[utf8]{inputenc} 
\usepackage[T1]{fontenc}    
\usepackage{amsfonts}       
\usepackage{nicefrac}       
\usepackage{amsmath}
\usepackage{amsthm}
\usepackage{multirow}
\usepackage{algorithm}
\usepackage{algorithmic}
\usepackage{float}
\usepackage{color}
\usepackage{enumitem}
\usepackage{bm}
\usepackage{wrapfig}
\usepackage{verbatim}
\usepackage[normalem]{ulem}
\usepackage{caption}
\usepackage{array}
\newcolumntype{L}[1]{>{\raggedright\let\newline\\\arraybackslash\hspace{0pt}}m{#1}}
\newcolumntype{C}[1]{>{\centering\let\newline  \\\arraybackslash\hspace{0pt}}m{#1}}
\newcolumntype{R}[1]{>{\raggedleft\let\newline \\\arraybackslash\hspace{0pt}}m{#1}}

\newtheorem{proposition}{Proposition}
\newtheorem{definition}{Definition}

\newcommand{\tcb}{\textcolor{blue}}

\def\R{\mathbb{R}}

\ifCLASSOPTIONcompsoc
  \usepackage[nocompress]{cite}
\else
  \usepackage{cite}
\fi

\ifCLASSINFOpdf
\else
\fi

\begin{document}
\title{Topology-aware Tensor Decomposition for Meta-graph Learning}

\author{\IEEEauthorblockN{Hansi~Yang}
	\IEEEauthorblockA{CSE HKUST\\
		hyangbw@cse.ust.hk}
	\and
	\IEEEauthorblockN{Peiyu~Zhang}
	\IEEEauthorblockA{EE, HUST\\
		u202012665@hust.edu.cn}
	\and
	\IEEEauthorblockN{Quanming~Yao}
	\IEEEauthorblockA{
		\textit{EE, Tsinghua}\\
		qyaoaa@tsinghua.edu.cn}}



\markboth{Journal of \LaTeX\ Class Files,~Vol.~14, No.~8, August~2015}%
{Shell \MakeLowercase{\textit{et al.}}: Bare Demo of IEEEtran.cls for Computer Society Journals}

\IEEEtitleabstractindextext{%
\begin{abstract}
Heterogeneous graphs generally refers to graphs with different types of nodes and edges. 
A common approach for extracting useful information from heterogeneous graphs 
is to use meta-graphs,
which can be seen as a special kind of directed acyclic graph (DAG)
with same node and edge types as the heterogeneous graph. 
However, 
how to design proper meta-graphs is challenging. 
Recently, 
there have been many works on learning suitable meta-graphs from a heterogeneous graph. 
Existing methods generally introduce continuous weights for edges that are independent of each other, 
which ignores the topological stucture of meta-graphs and can be ineffective. 
To address this issue, 
we propose a new viewpoint from tensor on learning meta-graphs.
Such a viewpoint not only helps interpret the limitation of existing works by 
CANDECOMP/PARAFAC (CP) decomposition,
but also inspires us to 
propose a topology-aware tensor decomposition, 
called \textit{TENSUS}, 
that reflects the structure of DAGs.
The proposed topology-aware tensor decomposition is easy to use
and simple to implement,
and it can be taken as a plug-in part to upgrade many existing works,
including node classification and recommendation 
on heterogeneous graphs.
Experimental results on different tasks demonstrate 
that the proposed method can significantly improve the state-of-the-arts for all these tasks.
\end{abstract}

}

\maketitle

\IEEEdisplaynontitleabstractindextext

\IEEEpeerreviewmaketitle

\section{Introduction}\label{sec:introduction}
Heterogeneous graphs~\cite{yang2020heterogeneous} refer to graphs where nodes and edges 
can have multiple types. 
Since it can model different types of relations, 
heterogeneous graph naturally appears in many real-world scenarios. 
An example is the academic network
(e.g. the Microsoft Academic Graph~\cite{wang2020microsoft}). 
The nodes here can have multiple types, e.g., ``paper'', ``author'' or ``institution'', 
and edges indicate different relations, e.g. 
an edge between an ``author'' node and a ``paper'' node 
indicates ``authorship'' relationship, 
while an edge between an ``author'' node and an ``institution'' node 
indicates ``affiliation'' relationship. 
Another example is a social network~\cite{gjoka-11}, 
which is also a heterogeneous graph 
if we allow two 
users (nodes) 
to have
different 
interactions
(edges). 

Due to the presence of multiple node types and edge types, 
a heterogeneous graph potentially contains a lot of semantic information 
\cite{sun2011metapath,sun2012mining,huang2016meta,zhao2017meta}.
An early attempt to
exploit such information
is by using
\textit{meta-paths}~\cite{sun2011metapath}.
A meta-path is a path with edge types 
from a heterogeneous graph. 
It can be seen as a generalization of edge 
and defines a composite relation between nodes. 
Thus, different meta-paths will define different composite relations 
and extract different semantic information from the heterogeneous graph. 
There are many methods utilizing meta-paths to extract information from heterogeneous graphs. 
For example, metapath2vec~\cite{metapath2vec2017} proposes to obtain embeddings 
for different nodes by performing random walks along meta-paths. 
HAN~\cite{wang2019heterogeneous} and MAGNN~\cite{fu2020magnn}
improve upon this idea and use a graph neural network (GNN)~\cite{kipf2016gcn} 
to obtain node embeddings. 

However, the path structure can be too restricted to define complex relations between nodes. 
To alleviate this problem,
\textit{meta-graphs}~\cite{huang2016meta,zhao2017meta} 
generalizes meta-path to more flexible structures other than path. 
A meta-graph 
is represented by a directed acyclic graph (DAG), 
whose nodes and edges have 
the same types as the heterogeneous graph. 
Similar to the meta-path, 
a meta-graph also define a composite relation 
between nodes in the heterogeneous graph. 
And since it uses a more flexible DAG structure, 
it can define more complex relations 
and help extract semantic information from the heterogeneous graph. 
Nevertheless, 
the meta-graphs
still need to be predefined,
which requires prior knowledge 
and huge human efforts to design.
Besides,
the performance of these methods 
are also sensitive to the choices of meta-graphs~\cite{zhao2017meta,ding2021diffmg}.

To eliminate human efforts in designing meta-graphs, 
recent works try to directly learn an informative meta-graph~\cite{NIPS2019_GTN, han2020gems,
ding2021diffmg}
from a given heterogeneous graph. 
Observed that
meta-graphs are 
derived from a plain DAG by assigning 
different types to edges.
Given a plain DAG, 
the meta-graph learning problem can be seen 
as learning the edge type for each edge in a given plain DAG. 
By assigning each edge a set of weights for its edge types, 
we can transform this problem to a continuous optimization problem.
This provides us a unified framework for these recent works. 
For example,
GTN~\cite{NIPS2019_GTN} considers a plain DAG with a path-like structure, 
and uses the same transformation introduced before to make it a continuous optimization problem.
HGT~\cite{hu2020hgt} computes mutual attention scores for different types of edges, 
which can be seen as a different way to optimize weights on edge types. 
GEMS~\cite{han2020gems} considers a general plain DAG than GTN 
and uses genetic search to find a proper meta-graph.
While GEMS can learn meta-graphs instead of only meta-paths like GTN,
its computational cost is much higher due to 
the genetic search algorithm for discrete optimization.
In recent years, neural network architecture search(NAS)~\cite{zoph2016neural} has demonstrated powerful capabilities in graphs~\cite{10.1145/3584945,zhang2023autogt}. 
Some NAS-based methods for hetergeneous graph have been researched and applied to learning tasks of heterogeneous graphs~\cite{gao2023hgnas++,ding2021diffmg} 
DiffMG~\cite{ding2021diffmg} is current the state-of-the-art on designing meta-graph.
By parameterization choices of edge types
as a differentiable propagation matrix in GNN,
it learns the meta-graph in an end-to-end  manner efficiently by stochastic gradient descent.

However, 
existing methods do not consider the effect of different DAG topological structures. 
For example, consider the two plain DAGs in Fig.~\ref{fig:mp2}, 
whose edges have no specific 
edge types. 
While they both have the same number of edges, 
the relations between their edges are definitely not the same and worth consideration. 
While
GTN simply cannot learn meta-graphs, 
the genetic search in GEMS 
and 
differentiable parameterization in DiffMG
both
ignore above topological difference.

Motivated by this limitation, 
we propose a new meta-graph learning method that is aware of 
the topological structures of different DAGs.  
We first view the meta-graph learning 
as a problem on tensor, and demonstrate that 
GTN/DiffMG can indeed be seen as using 
the simple rank-1 CP (CANDECOMP / PARAFAC) tensor decomposition
to express a tensor~\cite{Bro1997PARAFAC, tsurvey2009}, 
which only has limited expressive power. 
From the tensor perspective, 
we propose to introduce a family of tensor decompositions 
that can change with the topological structure of different DAGs. 
While a number of tensor decompositions~\cite{2016arXiv160900893C,2017arXiv170809165C} 
have been developed to incorporate structure information into tensor decompositions,
they are mainly designed for simple structures and not
suitable for learning meta-graphs, 
which can have very different structures. 
Moreover, existing applications of tensor decompositions 
mainly focus on the approximation of a given tensor, whereas here the goal is
to learn the tensor from data by imposing a tensor structure based on the
associated meta-graphs.
Extensive empirical results on different downstream tasks of heterogeneous graphs
show that it outperforms the state-of-the-arts.

Our contributions are summarized as:
\begin{itemize}[leftmargin=*]
\item We give a tensor view on meta-graph learning,
and highlight the drawbacks of existing methods from the perspective 
of tensor decomposition;

\item 
Based on tensor decomposition,
we propose a novel parameterization 
that is aware of the topological structures of different DAGs;

\item We conduct experiments on various downstream applications of heterogeneous graph 
as well as the logic rule learning problem on knowledge graph. 
Results demonstrate the performance gain of our topology-aware parameterization. 
\end{itemize}

\textbf{Notations.} We use boldface lowercase letters (e.g., $\vw$)
to denote vectors, 
boldface uppercase letters (e.g., $\mA$) to denote matrices, 
and calligraphed uppercase letter (e.g., $\gT$) to denote tensors. 
The $i$th element of a vector $\vw$ is denoted $\vw(i)$.
element $(i,j)$ of a matrix $\mA$ is denoted $\mA(i,j)$, 
and element $(i_1, \dots, i_k)$ of a tensor $\gT$ is denoted $\gT(i_1, \dots, i_k)$. 
We also use $:$ (e.g., $\mA(i,:)$ or $\gT(i_1, \dots, :, \dots, i_k)$)
to denote all elements in a matrix or tensor along a specific axis. 

\section{Background:
Heterogeneous Graphs}
\label{sec:hetegraph}

\begin{figure*}[ht]
	\centering
	\subfigure[Example heterogeneous graph. ]
	{\includegraphics[width=.35\textwidth]{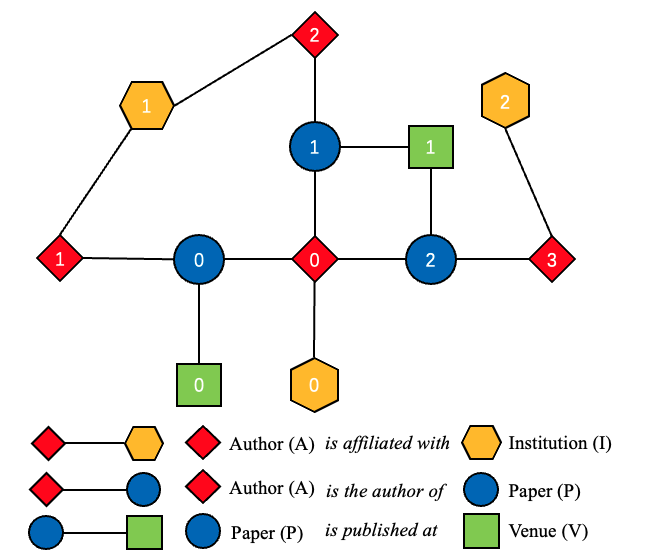}\label{fig:schema}}
	\qquad
	\begin{minipage}[b]{.4\textwidth}
		\subfigure[Meta-path. \label{fig:mp1}]{\includegraphics[width=\linewidth]{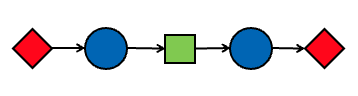}} \\
		\subfigure[Meta-graph. \label{fig:mg1}]{\includegraphics[width=\linewidth]{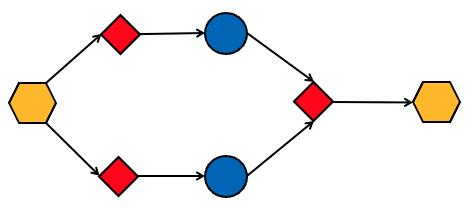}} 
	\end{minipage}
	\vspace{-10px}
	\caption{An example heterogeneous graph, and an associated meta-path and meta-graph.
		In this example, 
		the edge type is implicitly specified by the two node types that the edge
		connects. }
	\label{fig:hetg}
\end{figure*}

\begin{definition}[Heterogeneous Graph~\cite{yang2020heterogeneous}]
\label{def:hin}
A heterogeneous graph is a graph 
$\mathcal{G} = \{\mathcal{V}, \mathcal{E}, \gN, \mathcal{R}, f_{\gN}, f_{\mathcal{R}}\}$, 
where $\mathcal{V}$ is the set of nodes, $\mathcal{E}$ is the set of edges,
$\gN$ is the set of node types, $\gR$ is the set of edge types,
$f_{\gN}: \mathcal{V} \rightarrow \gN$ 
is a mapping 
from nodes
to node types,
and $f_{\mathcal{R}}$ is a mapping 
from edges
to edge types.
\end{definition}
When $|\gN|=| \mathcal{R} | =1$, 
a heterogeneous graph reduces to a homogeneous graph.  In this paper, we focus on
the case where both $|\gN|$ and
$| \mathcal{R} |$ are larger than 1. 
An example is shown in Fig.~\ref{fig:schema}, which has
4 types of nodes (author (A), paper (P), institution (I), venue (V)),  
and 6 types of edges
specifying the types of nodes connected (A-I/I-A, A-P/P-A, P-V/V-P).

Given a heterogeneous graph
with $C$ edge types, we have a set of  $C$ adjacency matrices $\sA = \{ \mA^1$, $\dots$, $\mA^C \}$, one for each edge type.
A naive approach is to ignore the edge types, which
reduces the
heterogeneous graph
to a homogeneous graph.
Standard graph data mining algorithms
(such as DeepWalk~\cite{perozi2014deepwalk}) 
or graph neural networks (GNN) (such as GCN~\cite{kipf2016gcn} or
GAT~\cite{gat2018}) 
can then be used.
However, edge types are often critical, and
ignoring them can lead to poor performance. 

To utilize edge type information,
a common approach 
is to use \textit{meta-paths}:
\begin{definition}[Meta-path~\cite{sun2011metapath}]
	\label{def:meta_path}
	A meta-path $\mathcal{P}$ is a sequence of node and edge types:
	$\gP = t_0 \! \xrightarrow{e_1} \! t_1 \! \xrightarrow{e_2} \! \dots \! \xrightarrow{e_M} \! t_{M}$,
	where $t_0, \dots, t_{M} \! \in \! \mathcal{N}$ and $e_1, \dots, e_M \! \in \! \mathcal{R}$.
\end{definition}
Fig.~\ref{fig:mp1}
shows an example meta-path ``A-P-V-P-A''.
Specifically, it expresses the relationship that two authors publish papers in the same
venue.
A natural and more powerful extension 
of the meta-path
is 
the \textit{meta-graph}, which
uses
a more flexible DAG structure.
\begin{definition}[Meta-graph~\cite{sun2012mining,huang2016meta}]
\label{def:meta_graph}
A meta-graph $\mathcal{M}$ is a directed acyclic
graph (DAG),
where its node types and edge types are subsets of $\gN$ and $\gR$, respectively.
\end{definition}
To be consistent with the meta-path,
we also restrict the meta-graph to have only a single source node 
and a single target node.
Fig.~\ref{fig:mg1} shows an example meta-graph, 
which describes a complex relation 
in which two authors 
from one institution
co-author papers with another author
in another institution. 

\subsection{Tensor Decomposition}
\label{sec:rel:tn}

A tensor is a high-dimensional array describing multilinear relations among objects.
Consider an $M$-order tensor 
$\mathcal{T}$ 
of size $C_1 \times \cdots \times C_M$.
As 
$M$ increases, 
obviously
it becomes more expensive to store all 
its entries. In recent decades, a number of 
tensor decompositions with
more efficient storage 
have been developed 
\cite{tsurvey2009,2016arXiv160900893C,htensor2010,ttrain2011}.
A classic example is the CP (CANDECOMP/PARAFAC) decomposition~\cite{Bro1997PARAFAC, tsurvey2009}. 
The $(i_1, i_2, \dots, i_M)$-th entry of $\mathcal{T}$ 
is given as:
\begin{equation} 
	\label{eq:cp}
	\mathcal{T}(i_1, \dots, i_M) 
	= \sum\nolimits_{r=1}^d \prod\nolimits_{m=1}^M \mG_m(i_m, r),
\end{equation}
where each 
$\mG_m\in\R^{C_m \times d}$, and
$d$ is the \textit{rank}.
Another well-known example is the
Tucker decomposition~\cite{tsurvey2009, Tucker1966Some}. 
However, its number of parameters 
still scales exponentially with the tensor order, and
can be problematic even for a moderate $M$. 

To avoid the curse of dimensionality
in high-order tensors, a class of
more complex tensor decompositions
called tensor networks
have been proposed
\cite{2016arXiv160900893C,2017arXiv170809165C}.
A popular
example is
the 
tensor train (TT)~\cite{ttrain2011}, 
which decomposes
$\mathcal{T}$
as:
\begin{align} 
\lefteqn{\mathcal{T}(i_1, \dots, i_M) =}\nonumber \\ 
& & \sum\nolimits_{r_1=1}^{d_1} \dots \sum\nolimits_{r_{M-1}=1}^{d_{M-1}}
\mW_1(i_1, r_1) \gW_2(i_2, r_1, r_2) \nonumber\\
&&\cdots 
\gW_{M-1}(i_{M-1}, r_{M-2}, r_{M-1}) 
\mW_M(i_M, r_{M-1}), \label{eq:tt1}
\end{align}
where $\mW_1 \in \R^{C_1 \times d_1}, \mW_M \in \R^{C_M \times d_{M-1}}$,
and $\gW_m$'s are 3-order tensors with size $C_m \times d_{m-1} \times d_m$. 
These matrices and tensors are called \textit{core tensors},
and $d_m$'s are the ranks. 
As the name implies, the TT has a train-like (sequential) computation process.

The TT decomposition has been successfully used in many applications 
\cite{ttrain2011,Novikov2014PuttingMO,novikov2015tensorizing}.
However, 
it assumes a simple line structure in the tensor, and is not suitable for more
complicated tensor
structures. 
Recently, Li and Sun~\cite{Li2020Evolutionary} attempt to directly search for
an appropriate tensor decomposition by genetic search. 
Nevertheless, this is very expensive.

\section{Related Works}

\begin{figure}[ht]
\centering
\includegraphics[width=.5\textwidth]{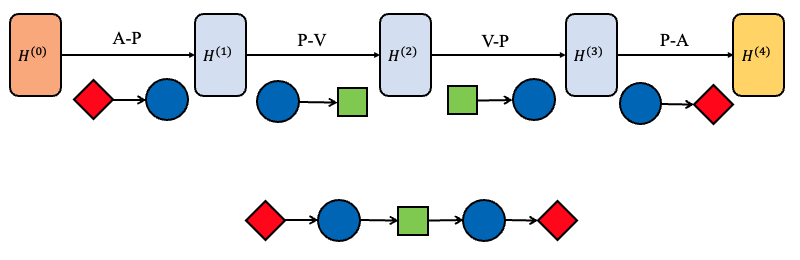}\label{fig:gnnmp}
	\vspace{-10px}
\caption{Computation of node embeddings (top) using a given meta-path (bottom).}
	\label{fig:gnns}
\end{figure}

\subsection{Learning with given meta-path / meta-graph}
\label{sec:given}

The edge types in 
the meta-path
can help
guide the GNN graph convolutions
and thus produce more informative node embeddings.
For example, 
MAGNN~\cite{fu2020magnn} 
updates
the node embedding 
$\mH^{(k)}$
at the $k$th iteration
as: 
\begin{equation} \label{eq:fixed}
\mH^{(k)} = f(\mH^{(k-1)}; \mA^{i_{k}}), \; k=1,2,\dots,K,
\end{equation} 
where
$K$ is the number of edges
in a given meta-path,
$\mH^{(0)}$ is the initial node embedding 
(usually set to the node feature matrix $\mX$), 
$f(\mH; \mA) = \sigma(\mA\mH\bm{\Theta})$ is the standard graph
convolution~\cite{kipf2016gcn} 
with
parameter 
$\bm{\Theta}$, 
and $\mA^{i_{k}}$ 
(with $i_k \in \{ 1, \dots, C \}$)
is
the adjacency matrix of the $k$th edge in the meta-path.
Fig.~\ref{fig:gnns} shows 
an example
with $K=4$.
Similarly, meta-graphs can also be used to help the GNN to learn node embeddings.


\begin{figure*}[ht]
	\centering
\subfigure[Computing node embeddings from meta-paths (left) and two example meta-paths (right).]
	{\includegraphics[width=.4\textwidth]{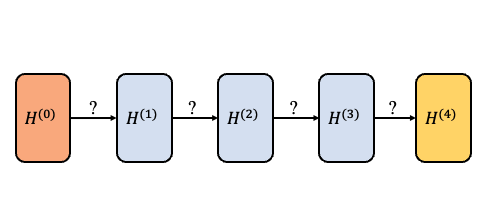}\label{fig:dagmp}}
	\subfigure[Computing node embeddings from meta-graphs (left) and two example meta-graphs (right).]
	{\includegraphics[width=.4\textwidth]{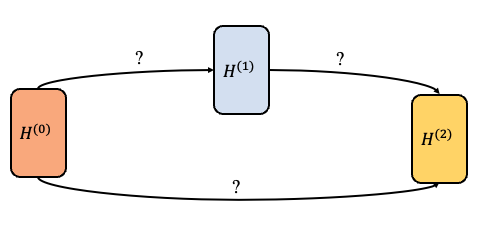}\label{fig:dagmg}}
	
	\vspace{-10px}
	\caption{Learning meta-path (a) and meta-graph (b) to compute node embeddings with edge types from the heterogeneous graph shown in Fig.~\ref{fig:schema}.}
	\label{fig:mp2}
\end{figure*}

\subsection{Meta-path / meta-graph learning }

Section~\ref{sec:given}
rely on predefined meta-paths
or meta-graphs. 
However,
designing
meta-paths and/or meta-graphs
require domain knowledge,
and can be expensive or even infeasible.
A natural extension is to directly learn the meta-path or meta-graph from the heterogeneous graph. 
Earlier works such as GTN~\cite{NIPS2019_GTN} and HGT~\cite{hu2020hgt} 
focus on the learning of meta-paths.  
Since
the edge type to be used in each segment of 
the meta-path is not known
(Fig.~\ref{fig:dagmp}),
the update in (\ref{eq:fixed}) is extended and
the node embedding
is obtained 
as
a weighted combination 
of the convolution outputs from all $C$ candidate edge types:
\begin{align}
\mH^{(k)} = 
\sum\nolimits_{i=1}^C \alpha^{(k-1, k)}_{i} f(
\mH^{(k-1)}; 
\mA^i), \;\; k=1,\dots,K,
\label{eq:mpb}
\end{align}
where 
$\{ \alpha^{(k-1, k)}_{i} \}_{i=1}^C$  
are the weights 
(which sum to 1) that
are learned together
with the graph convolution parameter $\bm{\Theta}$.
After convergence, 
for each edge $\mH^{(k-1)} \rightarrow \mH^{(k)}$,
the edge type $i$ with the largest weight $\alpha^{(k-1, k)}_{i}$ 
is chosen to form
the meta-path.

More recently, 
{\em meta-graph learning} methods
(such as GEMS~\cite{han2020gems} and DiffMG
\cite{ding2021diffmg})
learn meta-graphs with general structures (Fig.~\ref{fig:dagmg}).
here, we focus on 
the state-of-the-art
DiffMG.
At the $k$th iteration,
DiffMG 
updates
the node embedding by 
using 
node embeddings 
$\{\mH^{(0)}, \dots, \mH^{(k-1)} \}$
at all previous iterations: 
\begin{equation}
\mH^{(k)} \!\! =\!\!  
\sum\nolimits_{i=1}^C 
\sum\nolimits_{j=0}^{k-1} 
\alpha^{(j, k)}_{i} f(
\mH^{(j)}; 
\mA^i), \;\; k=1,\dots,K,
\label{eq:dmgb}
\end{equation}
where 
the learnable weights 
$\alpha^{(j, k)}_{i}$'s  satisfy
$\sum_{i=1}^C
\alpha^{(j, k)}_{i} =1$ for each $(j,k)$ pair.
After convergence, DiffMG also 
obtains a meta-graph by
choosing the edge type $i_{j,k}$ with the largest 
$\alpha^{(j, k)}_{i}$ in each 
$\mH^{(j)} \rightarrow \mH^{(k)}$
edge.

\begin{figure*}[t]
	\centering
	\vspace{-10px}
	\subfigure[GTN/DiffMG with $N=0$ (2 node embeddings). \label{fig:mot1}]
	{\includegraphics[width=0.28\textwidth]{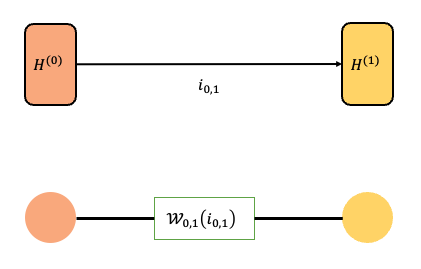}}
	\subfigure[GTN with 4 node embeddings. \label{fig:mot3}]
	{\includegraphics[width=0.4\textwidth]{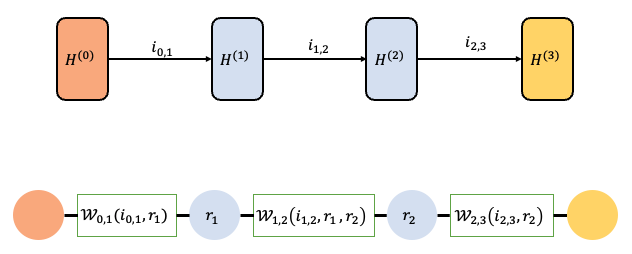}}
	\subfigure[DiffMG with $N=1$ (3 node embeddings).  \label{fig:mot2}]
	{\includegraphics[width=0.28\textwidth]{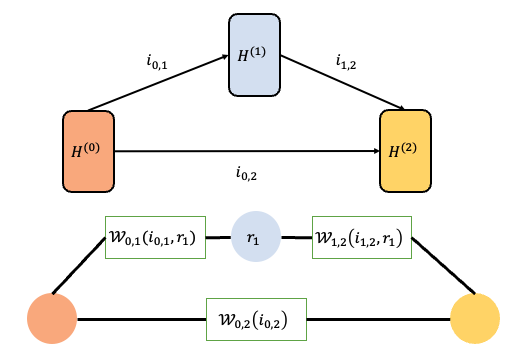}}
	\vspace{-10px}
	\caption{The computation graph (top) for node embeddings and corresponding
	tensor decomposition (bottom).}
	\label{fig:tn}
\end{figure*}

\section{Tensor Formulation for Meta-Graph Learning}
\label{ssec:tform}

In this section, we 
formulate
meta-graph learning 
in terms of tensors. 
Section~\ref{ssec:rev} first shows the connection between some existing works on
meta-path/meta-graph learning
with rank-1 CP decomposition. 
Section~\ref{sec:encode} then proposes a novel tensor decomposition for meta-path/meta-graph learning. 
Section~\ref{sec:integrate} introduces how the proposed decomposition is used for meta-graph learning. 

\subsection{Existing Meta-Graph Learning as Rank-1 CP Decomposition}
\label{ssec:rev}

Since each $\mH^{(k)}$ in
(\ref{eq:dmgb})
is computed from $\{\mH^{(0)}, \dots, \mH^{(k-1)} \}$, 
the computation graph for $\{\mH^{(0)}, \dots, \mH^{(K)}\}$
has $M \equiv 
K(K+1)/2$ edges.
We index these $M$ edges as
$\{ g_1, \dots, g_M \}$, in which each $g_m=(j,k)$
corresponds to the edge $\mH^{(j)} \rightarrow \mH^{(k)}$.
There are 
thus
a total of $C^M$ candidate meta-graphs.
Each such candidate 
can be specified by the $M$-tuple 
$(i_{g_1}, 
i_{g_2}, 
\dots, i_{g_M})$, 
where $i_{g_m}\equiv i_{j,k} \in \{1, \dots, C\}$ represents the edge type used for
edge $g_m$ connecting
$\mH^{(j)}$ to 
$\mH^{(k)}$.
As noted above,
existing methods (such as GTN, HGT and DiffMG) 
form the meta-path/graph by
selecting the edge type 
$i_{j,k}$
with the largest weight 
for each $(j,k)$ pair in 
(\ref{eq:dmgb}).
This is equivalent to  finding
the meta-graph $\mathcal{M}$ which maximizes the score
\[ w_{\gM} = \prod\nolimits_{k=1}^{K} \prod\nolimits_{j=0}^{k-1} \alpha^{(j,
k)}_{i_{j,k}} \]
(for 
GTN~\cite{NIPS2019_GTN}, this reduces to the form $w_{\gM} =
\prod\nolimits_{k=1}^{K}  \alpha^{(k-1, k)}_{i_{k}}$). 
A compact way to store the scores of all candidate
meta-graphs is to use a $M$-order tensor $\gT$, with:
\begin{equation} \label{eq:diffcp}
\gT(i_{g_1}, \dots, i_{g_M}) = 
\prod\nolimits_{k=1}^{K} \prod\nolimits_{j=0}^{k-1} \alpha^{(j, k)}_{i_{j,k}}. 
\end{equation} 
Note that this is a rank-1 CP decomposition in (\ref{eq:cp}), with $d=1$ and 
$\mG_m(i_{j,k}, 1) = \alpha^{(j, k)}_{i_{j,k}}$.  
In other words, DiffMG is implicitly using a rank-1 CP decomposition.
However, it is known that
rank-1 CP decomposition only has limited expressive power
\cite{tsurvey2009}.

\subsection{Encoding Meta-Path/Meta-Graph Structure by Tensoer Decomposition}
\label{sec:encode}

In this section, we propose a novel tensor decomposition, called
\textit{TENSUS}
(Tensor Meta-graph Learning).
It is more powerful than
the rank-1 CP decomposition,
and
performs meta-path/meta-graph
learning by 
taking the local graph structures 
into consideration.

\subsubsection{Encoding Meta-Paths}
\label{sec:metapath}

We first consider the special case of meta-path learning. 
The simplest meta-path involves only two node embeddings: $\mH^{(0)}$ and $\mH^{(1)}$ 
(Fig.~\ref{fig:mot1}). 
As there is only one edge
($\mH^{(0)} \rightarrow \mH^{(1)}$),
the weights for all $C$ candidate edge types  
can be stored
in a 
vector 
$\gW_{0,1} \in \R^C$, 
where the subscripts correspond to the 
node embedding
indices.

Next, consider the case with 
$3$ node embeddings $\mH^{(0)}, \mH^{(1)}$ and $\mH^{(2)}$, in which
$\mH^{(1)}$ 
is computed
from $\mH^{(0)}$ and then 
$\mH^{(2)}$ 
computed
from $\mH^{(1)}$. 
Recall from (\ref{eq:diffcp}) that $\gT(i_{0,1}, i_{1,2}) = \alpha^{(0, 1)}_{i_{0,1}} \alpha^{(1, 2)}_{i_{1,2}}$. 
The tensor $\gT$ (which reduces to a 
$C\times C$ 
matrix here) thus equals
$\bm{\alpha}^{(0, 1)}(\bm{\alpha}^{(1, 2)})^\top$ (where
$\bm{\alpha}^{(0, 1)}=
[\alpha^{(0, 1)}_i]$ 
and $\bm{\alpha}^{(1, 2)}=
[\alpha^{(1, 2)}_i]$)
and is
rank-1. 
In the following, we propose to replace this by a rank-$d$ matrix, 
which is more expressive 
than a rank-1 matrix but less expensive than a full matrix:
\begin{align}
\gT(i_{0,1}, i_{1,2}) = \sum_{r_1=1}^d \mW_{0,1}(i_{0,1}, r_1)
\mW_{1,2}(i_{1,2}, r_1),
\label{eq:gtn1}
\end{align}
where 
$\mW_{0,1}, \mW_{1,2} \in \R^{C \times d}$
are 
learnable matrices corresponding to the edges
$\mH^{(0)} \rightarrow \mH^{(1)}$ and
$\mH^{(1)} \rightarrow \mH^{(2)}$, respectively. 

When there are $4$ node embeddings, $\gT$ becomes a $3$-order tensor.
We propose to use the tensor train (TT) decomposition
(Section~\ref{sec:rel:tn})
which follows a path-like computation similar to meta-path. 
Using (\ref{eq:tt1}),
the tensor $\gT$ is 
(Fig.~\ref{fig:mot3}):
\begin{align}
\lefteqn{\mathcal{T}(i_{0,1},  i_{1,2}, i_{2,3})} \nonumber\\
&	\!\!\! = & \!\!\!\!\!\!  \sum_{r_{1}=1}^{d} \!\!\! \sum_{r_{2}=1}^{d} \mW_{0,1}(i_{0,1}, r_{1})
\gW_{1,2}(i_{1,2}, r_{1}, r_{2}) \mW_{2,3}(i_{2,3}, r_{2}). \label{eq:tmp3}
\end{align}
Similar to (\ref{eq:gtn1}),
each $\gW_{j,k}$ in (\ref{eq:tmp3}) corresponds to the edge $\mH^{(j)} \rightarrow \mH^{(k)}$,
and each index $r_k$ corresponds to an intermediate $\mH^{(k)}$.
When $d=1$, (\ref{eq:tmp3}) reduces to
$\gT(i_{0,1}, i_{1,2}, i_{2,3}) = 
\mW_{0,1}(i_{0,1}, 1) \gW_{1,2}(i_{1,2}, 1, 1) \mW_{2,3}(i_{2,3}, 1)$,
which is equivalent to 
$\alpha^{(0, 1)}_{i_{0,1}} \alpha^{(1, 2)}_{i_{1,2}} \alpha^{(2, 3)}_{i_{2,3}}$
in
(\ref{eq:diffcp}), 
with only a difference in notations. 

For the general case with node embeddings $\mH^{(0)}, \dots,\mH^{(K)}$,
we again use the TT decomposition on higher-order tensors. 
Similar to (\ref{eq:tmp3}), we  have:
\begin{align}
\mathcal{T}& (i_{0,1},  \dots, i_{K-1,K}) = \sum_{r_{1}=1}^{d} \dots \sum_{r_{K-1}=1}^{d} \nonumber \\
& \mW_{0,1}(i_{0,1}, r_{1}) \gW_{1,2}(i_{1,2}, r_{1}, r_{2}) \dots \mW_{K-1,K}(i_{K-1,K}, r_{K-1}). 
\label{eq:tdp}
\end{align}
When $d=1$,
this also becomes equivalent to (\ref{eq:diffcp}).

\subsubsection{Encoding Meta-Graphs}

In this section, we consider the encoding of meta-graphs. 
When there are only 2 nodes, the meta-graph is the same as a meta-path
as we only have one edge. 
When there are $3$ nodes, 
recall from (\ref{eq:dmgb}) that $\mH^{(2)}$ is computed by
summing the graph convolutions due to $\mH^{(0)}$ and $\mH^{(1)}$
(Fig.~\ref{fig:mot2}).
Let $\gT_1(i_{0,1}, i_{1,2})$ and $\gT_2(i_{0,2})$ be the matrix and vector
containing the scores for $\mH^{(0)} \rightarrow \mH^{(1)} \rightarrow \mH^{(2)}$ and $\mH^{(0)} \rightarrow \mH^{(2)}$, respectively.
Intuitively, as the two branches 
$\mH^{(0)} \rightarrow \mH^{(1)} \rightarrow \mH^{(2)}$ and
$\mH^{(0)} \rightarrow \mH^{(2)}$
are independent, 
their scores
can be simply multiplied together to obtain the score of the whole graph as:
\begin{equation} \label{eq:tmp}
\gT(i_{0,1}, i_{0,2}, i_{1,2}) = \gT_1(i_{0,1}, i_{1,2}) \gT_2(i_{0,2}).
\end{equation} 
To represent all possible 
$(i_{0,1}, i_{0,2}, i_{1,2})$
combinations,
$\gT_{1}$ can be stored as a $C\times C$ matrix,  and
$\gT_{2}$ as a $C$-dimensional vector.
Similar to (\ref{eq:gtn1}),
we propose to express $\gT_{1}$ by rank-$d$ decomposition, i.e.,
$$\gT_{1}(i_{0,1}, i_{1,2})  = \sum_{r_1=1}^d \mW_{0,1}(i_{0,1}, r_1) \mW_{1,2}(i_{1,2}, r_1), $$ 
with
$\mW_{0,1}, \mW_{1,2} \in \R^{C \times d}$.
Thus,
(\ref{eq:tmp})
becomes:
\begin{equation}
\gT(i_{0,1}, i_{0,2}, i_{1,2}) 
= 
\sum_{r_1=1}^d 
\! \mW_{0,1}(i_{0,1}, r_1) \mW_{1,2}(i_{1,2}, r_1) \vw_{0,2}(i_{0,2}).
\label{eq:tne}
\end{equation}
As in section~\ref{sec:metapath}, when $d=1$, (\ref{eq:tmp}) reduces to
$\gT(i_{0,1}, i_{1,2}, i_{2,3}) = 
\mW_{0,1}(i_{0,1}, 1) \mW_{1,2}(i_{1,2}, 1) \vw_{0, 2}(i_{0, 2})$,
which is equivalent to 
$\alpha^{(0, 1)}_{i_{0,1}} \alpha^{(1, 2)}_{i_{1,2}} \alpha^{(0, 2)}_{i_{0,2}}$
in
(\ref{eq:diffcp}), 
with only a difference in notations. 


Fig.~\ref{fig:mot2}
shows
the computation in (\ref{eq:tne}) graphically.
$\mW_{0,1}$ and $\mW_{1,2}$  in
(\ref{eq:tne}) 
share an index 
$r_1$
as the 
edges 
correspond  to $\mH^{(0)} \rightarrow \mH^{(1)}$ and $\mH^{(1)} \rightarrow \mH^{(2)}$
share an internal node. 
Thus, index  
$r_1$ can be viewed as corresponding to 
the intermediate 
$\mH^{(1)}$,
as in (\ref{eq:tdp}).



For the general case with node embeddings $\mH^{(0)}, \dots,\mH^{(K)}$, 
we can similarly extend the correspondence in (\ref{eq:tdp}) and (\ref{eq:tne}) 
by introducing indices $r_k$ for node embedding $\mH^{(k)}$, 
and core tensor $\gW_{j,k}$ for the edge $\mH^{(j)} \rightarrow \mH^{(k)}$. 
Since the computation for $\mH^{(K)}$ is also different now,
$\gT$ also takes a different form and becomes: 
\begin{align}
\lefteqn{\mathcal{T}(i_{0,1}, \dots, i_{K-1, K})
= \sum\nolimits_{r_{1}=1}^{d}\cdots\sum\nolimits_{r_{K-1}=1}^{d}} \notag \\
& & \mW_{0,1}(i_{0,1}, r_{1}) \cdots \mW_{0,K-1}(i_{0,K-1}, r_{K-1}) \notag \\
& & \gW_{1,2}(i_{1,2}, r_1, r_2) \cdots \gW_{K-2, K-1}(i_{K-2, K-1}, r_{K-2}, r_{K-1}) \notag \\
& & \vw_{0,K}(i_{0,K}) \mW_{1,K}(i_{1,K}, r_{1})  \cdots \mW_{K-1,K}(i_{K-1,K}, r_{K-1}). 
\label{eq:tn0}
\end{align}
By keeping only the edges $\mH^{(0)} \rightarrow \mH^{(1)}, \mH^{(1)} \rightarrow
\mH^{(2)}, \dots, \mH^{(K-1)} \rightarrow \mH^{(K)}$, the above is reduced to
\begin{align*}
& \mathcal{T}(i_{0,1}, \dots, i_{K-1, K})
= \sum_{r_{1}=1}^{d} \dots \sum_{r_{K-1}=1}^{d} \\
& \mW_{0,1}(i_{0,1}, r_{1}) \gW_{1,2}(i_{1,2}, r_{1}, r_{2}) \dots \mW_{K-1,K}(i_{K-1,K}, r_{K-1}),
\end{align*}
which is of the same form in (\ref{eq:tdp}). 


\subsection{Optimization Algorithm}
\label{sec:integrate}

To obtain
the final node embedding
$\mH^{(K)}$, we
solve the following optimization problem:
\begin{align}
\min_{\bm{\Theta}, \{ \gW_{j,k} \} } & \Ls(\mH^{(K)}) \label{eq:opt}\\
\text{s.t. } & \sum\nolimits_{i_{g_1}=1}^C 
\cdots
\sum\nolimits_{i_{g_M}=1}^C 
\gT(i_{g_1}, \dots, i_{g_M}) = 1, \label{eq:constr1}\\
& \gT(i_{g_1}, \dots, i_{g_M}) \ge 0,  \label{eq:constr2}
\end{align}
where $\bm{\Theta}$ is the GNN model parameter,
and 
$\Ls$ is 
an appropriate loss function.
The constraints ensure that the scores in $\gT$
are non-negative and sum to $1$.

\subsubsection{Differentiable transformation}
\label{sec:xform}

To enforce the constraints (\ref{eq:constr1}) and (\ref{eq:constr2}),  
DiffMG uses the softmax trick. 
However,
this cannot be directly used here,
as 
there are additional indices $r_j$'s
in (\ref{eq:tn0}).
we re-formulate the constraints by re-parametrizing
$\gW_{j,k}$'s as:
\begin{align}
\gW_{j, k}(:, r_j, r_{k}) & = \text{softmax}(\gB_{j, k}(:, r_j, r_{k})), 
\label{eq:res1} \\
\gW_{0,k}(:, r_k) & = \text{softmax}(\gB_{0, k}(:, r_k)), 
\label{eq:res2} \\
\gW_{j, K}(:, r_j) & = \text{softmax}(\gB_{j, K}(:, r_j)), 
\label{eq:res3} \\
\gW_{0, K}(:) & = \text{softmax}(\gB_{0, K}(:)),
\label{eq:res} 
\end{align}
where the
learnable 
$\gB$'s have the same shapes as the corresponding
$\gW$'s.
The softmax operation 
is used
only along the first dimension of $\gB$. 
For example, for $\gW_{j, k}$
in (\ref{eq:res1}),
the softmax operation is:
\begin{align*}
\gW_{j, k}(i, r_j, r_{k}) = \frac{\exp(\gB_{j, k}(i, r_j, r_{k}))}{\sum_{l=1}^C \exp(\gB_{j, k}(l, r_j, r_{k}))}.
\end{align*}

The following proposition shows that 
all the 
tensor elements  are non-negative and
sum
to a constant.
\begin{proposition}
\label{pr:norm}
For 
$\gT$ in (\ref{eq:tn0}) with $\gW_{j,k}$'s in (\ref{eq:res1})-(\ref{eq:res3}), 
we have
$\gT(i_{g_1}$, $\dots$, $i_{g_M}) \ge 0$, 
and 
$\sum\nolimits_{i_{g_1}=1}^C 
\dots 
\sum\nolimits_{i_{g_M}=1}^C \gT(i_{g_1}$, 
$\dots$, 
$i_{g_M}) = d^{K-1}$.
\end{proposition}
Thus, 
to satisfy the constraints in (\ref{eq:constr1}) and
(\ref{eq:constr2}),
we only need to divide (\ref{eq:tn0}) by the constant $d^{K-1}$, as
follows:
\begin{align} 
\lefteqn{\mathcal{T}(i_{0,1}, \dots, i_{K-1, K})
	=\frac{1}{d^{K-1}} \sum\nolimits_{r_{1}=1}^{d}\cdots\sum\nolimits_{r_{K-1}=1}^{d}} \notag \\
& & \mW_{0,1}(i_{0,1}, r_{1}) \cdots \mW_{0,K-1}(i_{0,K-1}, r_{K-1}) \notag \\
& & \gW_{1,2}(i_{1,2}, r_1, r_2) \cdots \gW_{K-2, K-1}(i_{K-2, K-1}, r_{K-2}, r_{K-1}) \notag \\
& & \vw_{0,K}(i_{0,K}) \mW_{1,K}(i_{1,K}, r_{1})  \cdots \mW_{K-1,K}(i_{K-1,K}, r_{K-1}). 
\label{eq:tnc}
\end{align}

\subsubsection{Gradient-based Algorithm}
\label{sec:alg}


For DiffMG, similar to (\ref{eq:dmgb}), 
the node embeddings $\mH^{(K)}$ are iteratively computed as: 
\begin{align*}
\mH^{(k)}_{r_k}\!\! = & \sum\nolimits_{i=1}^C \gW_{0,k}(i, r_k) f(\mH^{(0)}, \mA^i) \\
&\!\! + \!\! \sum\nolimits_{0 < j < k} \frac{1}{d}\! \sum\nolimits_{r_{j}=1}^d \! \sum\nolimits_{i=1}^C \gW_{j,k}(i,
r_{j}, r_k) f(\mH^{(j)}_{r_{j}}, \mA^i),\\
\mH^{(K)}\!\! = &  \sum\nolimits_{i=1}^C \gW_{0,K}(i) f(\mH^{(0)}, \mA^i) \\
&\!\! + \!\! \sum\nolimits_{0 < j < K} \frac{1}{d}\! \sum\nolimits_{r_j=1}^d \! \sum\nolimits_{i=1}^C \gW_{j,K}(i,
r_{j}) f(\mH^{(j)}_{r_{j}}, \mA^i), 
\end{align*}
where the subscript $r_k$ in $\mH^{(k)}_{r_k}$ is the same as the index $r_k$ in
the decomposition (\ref{eq:tn0}). 
We can see that this reduces to DiffMG~\cite{ding2021diffmg} if we have $d=1$, 
and the decomposition in (\ref{eq:tn0}) can also reduce to the rank-1 CP decomposition.  
A higher-rank CP decomposition  can also have more expressive power, because tensors networks having higher ranks can  restore high-order tensors more precisely.
However, this generalization overlooks how we compute the node embedding $\mH^{(k)}$'s, 
and leads to worse results that will be demonstrated in experiments \ref{ssec:hyp}. 
These weights $\gW$'s, which are given by learnable parameters $\gB$'s, 
are simultaneously optimized with the convolutional parameters $\bm{\Theta}$. 

\subsubsection{Space and Time Complexities}

The decomposition in (\ref{eq:tnc}) contains
at most $CMd^2$ parameters. 
If we do not set $d$ too large, 
this will be negligible compared with the number of parameters in other parts of
the model
(e.g., $\bm{\Theta}$ in the graph convolution part), which can easily have thousands of parameters. 

For the time complexity, if we assume computing
$\bar{f}_{j,k}$ (resp. $\bar{f}_{j,k, r_j, r_k}$) takes unit time, 
then the time complexity should be  $\gO(N^2d^2)$ for 
TENSUS and
$\gO(N^2)$ for DiffMG where $\bar{f}$ is message passing functions like GCN~\cite{kipf2016gcn}. 
We will also compare the impact of different $d$'s on the performance 
and computational cost in Section~\ref{ssec:hyp}.

\begin{table*}[ht]
	\caption{Macro F1 scores on node classification for different heterogeneous graphs. (results of baseline were copied from DiffMG~\cite{ding2021diffmg})}
	\centering
	\small
	\begin{tabular}{c | c c c c c c c c | c c}
		\toprule
		     & metapath2vec* & GCN*   & GAT*   & HAN*  & MAGNN* & GTN  & HGT* & DiffMG* & TENSUS   \\ \midrule
		DBLP & 85.53        & 87.30 & 93.71 & 92.83 & 92.81 & 93.98 & 93.67 & 94.45 & \textbf{95.92} \\
		ACM  & 87.61        & 91.60 & 92.33 & 90.96 & 91.15 & 91.89 & 91.83 & 92.65 & \textbf{94.02} \\
		IMDB & 35.21        & 56.89 & 58.14 & 56.77 & 57.13 & 59.68 & 59.35 & 61.04 & \textbf{64.16} \\ \bottomrule
	\end{tabular}
	\label{tab:gtn}
\end{table*}

\begin{figure*}[ht]
	\centering
	\subfigure[IMDB for node classification. \label{fig:imdb}]
	{\includegraphics[width=0.3\textwidth]{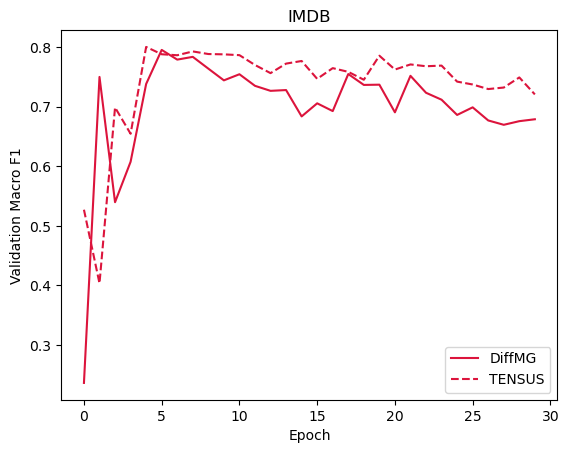}}
	\subfigure[Douban for recommendation. \label{fig:db} ]
	{\includegraphics[width=0.3\textwidth]{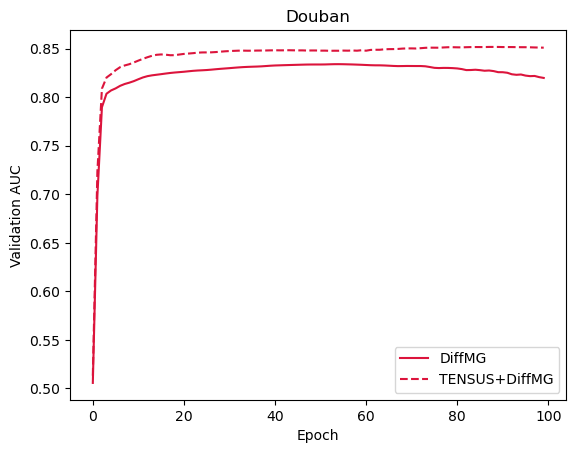}}
	\caption{Learning curves of different methods. }
\end{figure*}

\begin{table*}[ht]
	\caption{AUC (\%) on the recommendation task on different heterogeneous graphs. 
		(results of baseline were copied from DiffMG~\cite{ding2021diffmg}) }
	\centering
	\small
	\begin{tabular}{c | c c c c c c c c c | c c }
		\toprule
		& metapath2vec* & GCN*   & GAT*   & HAN*  & MAGNN* & GEMS* & GTN* & HGT* & DiffMG*  & TENSUS        \\ \midrule
		Amazon    &  58.17   & 66.64 & 55.70 & 67.35 & 68.26 & 70.66& 71.82 & 74.75 & 75.28 & \textbf{77.72} \\
		Yelp  &  51.98  & 58.98 & 56.55 & 64.28 & 64.73 & 65.12 & 66.27 & 68.07 & 68.77 & \textbf{70.38} \\
		Douban  & 51.60 & 77.95 & 77.58 & 82.65 & 82.44 & 83.00 & 83.26 & 83.38 & 83.78 & \textbf{84.44} \\ \bottomrule
	\end{tabular}
	\label{tab:rec}
\end{table*}

\section{Experiments}
\label{sec:expt}

In this section, we empirically verify the improvement 
of our proposed topology-aware parameterization. 
We integrate our parameterization with existing methods, 
as is introduced in Section~\ref{sec:integrate}. 
For the proposed method, we set $d=2$ unless otherwise specified. 
Experiments are implemented by PyTorch 
and run on a machine with a single NVIDIA RTX 2080 Ti GPU.

\subsection{Node Classification}
\label{sec:node}

Experiments are performed on three standard benchmark 
data sets 
\cite{wang2019heterogeneous, NIPS2019_GTN}:
\href{https://dblp.org/xml/}{DBLP},\href{https://paperswithcode.com/dataset/acm}{ACM} and \href{https://www.kaggle.com/datasets/lakshmi25npathi/imdb-dataset-of-50k-movie-reviews}{IMDB}
(Table~\ref{tab:hind}). 
The DBLP dataset has three types of nodes: paper (P), author
(A), and conference (C) where authors are labeled by their research areas;
the ACM dataset has nodes: paper
(P), author (A), and subject (S) where papers are labeled by research areas; and
the IMDB dataset has nodes: movies
(M), actors (A), and directors (D) where movies are
labeled by genres. 
The
nodes use the bag-of-words representation as input features.

\begin{table}[ht]
	\caption{Statistics of the graph datasets for node classification.}
	\centering
	\vspace{-10px}
	\begin{tabular}{c c c c}
		\toprule
	& DBLP & ACM  & IMDB  \\ \midrule
		\# nodes &18405    & 8994     & 12772    \\
		 \# edges & 67946    & 25922    & 37288  \\
		  \# edge types & 4 & 4 & 4  \\
		  \# features & 334         & 1902        & 1256  \\
		  \# training & 800         & 600         & 300    \\
		   \# validation & 400           & 300           & 300    \\
		     \# test & 2857  &  2125 & 2339 \\ \bottomrule
	\end{tabular}
	\label{tab:hind}
\end{table}

We compare our proposed method TENSUS with DiffMG~\cite{ding2021diffmg} as well as
the following baselines in~\cite{ding2021diffmg}: 
metapath2vec~\cite{metapath2vec2017},
GCN~\cite{kipf2016gcn},
GAT~\cite{gat2018},
HAN~\cite{wang2019heterogeneous}, 
MAGNN~\cite{fu2020magnn}, GTN~\cite{NIPS2019_GTN}
and HGT~\cite{hu2020hgt}. 
The experimental setting follows GTN~\cite{NIPS2019_GTN} and DiffMG~\cite{ding2021diffmg}. 
These baselines are selected by DiffMG~\cite{ding2021diffmg}, and we follow the setup of this paper. 
The embedding dimension is 64.
We use the Adam optimizer~\cite{adam} 
with its hyper-parameters (learning rate, weight decay and input dropout) 
tuned according to the validation set.
For performance evaluation, we use the average 
macro-F1 score
over 5 runs
with different random seeds. 

Table~\ref{tab:gtn} shows the macro-F1 scores.
Note that HAN and MAGNN, which rely on manually-designed meta-paths, 
do not have good performance as compared to GTN, HGT, and DiffMG. 
This indicates that manually designed meta-graphs can be limited in mining
task-dependent semantic information. 
DiffMG consistently achieves better performance than GTN and HGT 
due to the use of meta-graphs with flexible topologies, 
and TENSUS 
achieves the best performance 
among all these methods 
by encoding the topological structure of meta-graphs into parametrization.

Fig.~\ref{fig:imdb}
shows the learning curves
on the IMDB dataset.
As can be seen,
meta-graphs learned by 
TENSUS lead to 
better performance
than the baseline method.

\subsection{Link Prediction}


In this experiment,
we consider three commonly used recommendation datasets: 
Yelp, Douban movie,
and Amazon
(Table~\ref{tab:recd}). 
The Yelp dataset is a platform where users review businesses,
Douban is a social media community where users share reviews about movies. 
Amazon is a large e-commerce platform which contains users’ ratings for items. 

\begin{table}[ht]
	\caption{Statistics of the graph datasets for recommendation task. Boldface indicates the target edge type we want to predict. }
	\centering
	\vspace{-5px}
	\begin{tabular}{c c c c c}
		\toprule
	dataset & relation (A-B) & \# A  & \# B & \# A-B \\ \midrule
		& \textbf{User-Business (U-B)} & \textbf{16239}  & \textbf{14284}  & \textbf{198397}  \\
		 & User-User (U-U)  & 16239  & 16239 & 158590 \\
		 Yelp  & User-Compliment (U-Co) & 16239 & 11 & 76875 \\
		  & Business-City (B-C)  & 14284     & 47 & 14267 \\
		 & Business-Category (B-Ca)    & 14284        & 511 & 40009   \\ \midrule
		 \multirow{6}{*}{Douban} &\textbf{User-Movie (U-M)} & \textbf{13367} & \textbf{12677} & \textbf{1068278}  \\
		 & User-Group (U-G) & 13367 & 2753 & 570047 \\
		 & User-User (U-U)  & 13367  & 13367 & 4085 \\
		  & Movie-Actor (M-A)  & 12677   & 6311 & 33587 \\
		  & Movie-Director (M-D)    & 12677    & 2449 & 11276   \\
		 & Movie-Type (M-T)    & 12677   & 38 & 27668 \\ \midrule
		\multirow{4}{*}{Amazon} &\textbf{User-Item (U-I)} & \textbf{6170} & \textbf{2753} & \textbf{195791}  \\
		 & Item-View (I-V)  & 2753 & 3857 & 5694 \\
		  & Item-Category (I-C) & 2753 & 22 & 5508 \\
		  & Item-Brand (I-B)  & 2753   & 334 & 2753  \\ \bottomrule
	\end{tabular}
	\label{tab:recd}
\end{table}

\begin{figure*}[ht]
	\centering
	\vspace{-10px}
	\subfigure[DiffMG. \label{fig:dmgs}]
	{\includegraphics[width=0.46\textwidth]{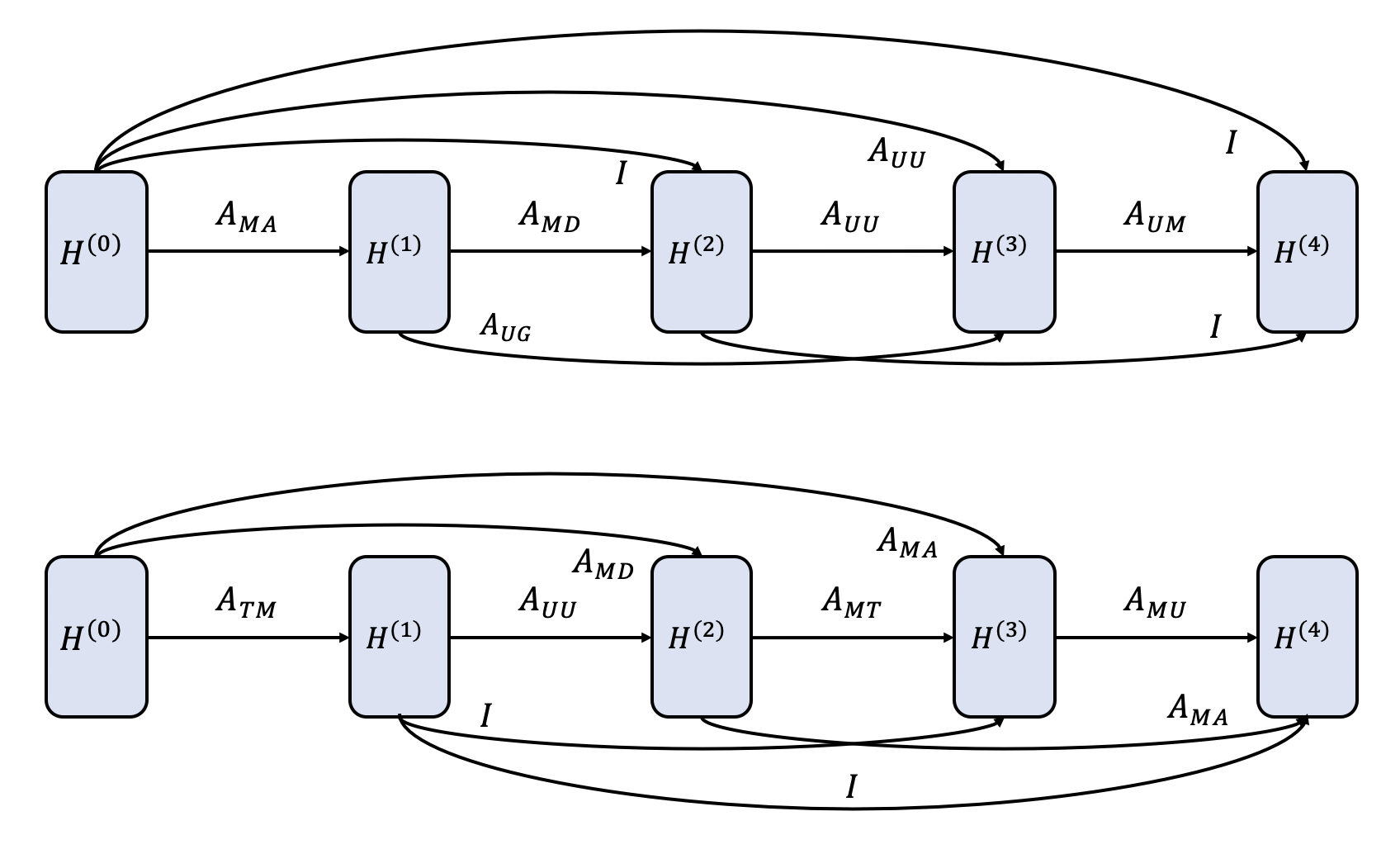}}
	\subfigure[TENSUS. \label{fig:tens}]
	{\includegraphics[width=0.46\textwidth]{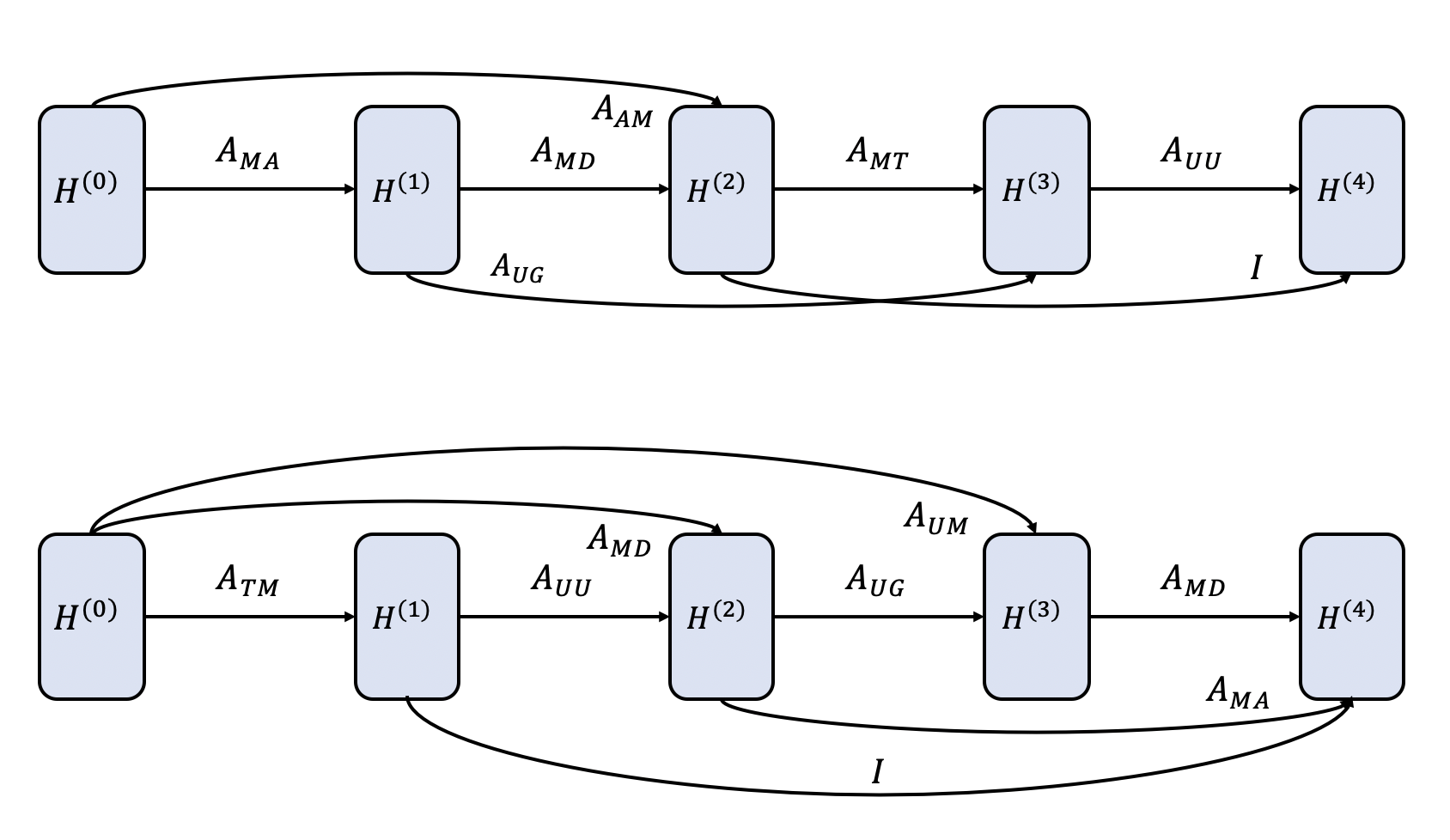}}
	\vspace{-8px}
	\caption{Meta-graphs learned by DiffMG and TENSUS for Douban dataset. }
	\vspace{-10px}
\end{figure*}

\begin{figure*}[ht]
	\centering
	\subfigure[Validation AUC vs rank $d$.  \label{fig:rank}]{\includegraphics[width=0.48\columnwidth]{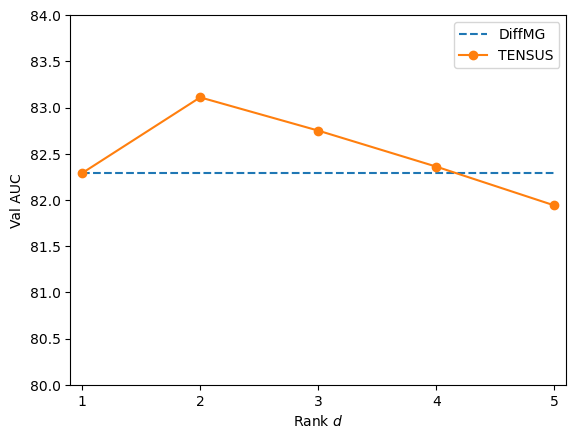}}
	\subfigure[Test AUC vs rank $d$.  \label{fig:rankt}]{\includegraphics[width=0.48\columnwidth]{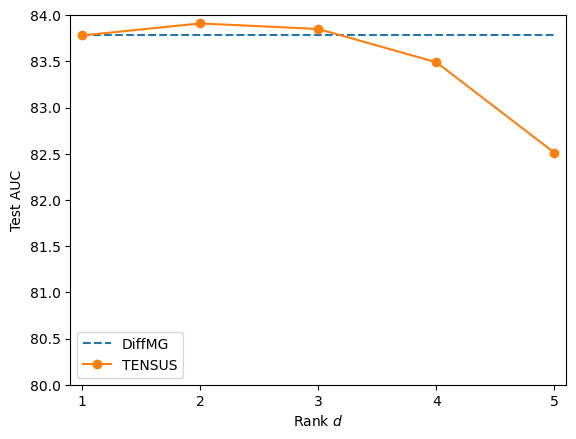}}
	\subfigure[Training time (min) vs rank $d$. \label{fig:time}]{\includegraphics[width=0.48\columnwidth]{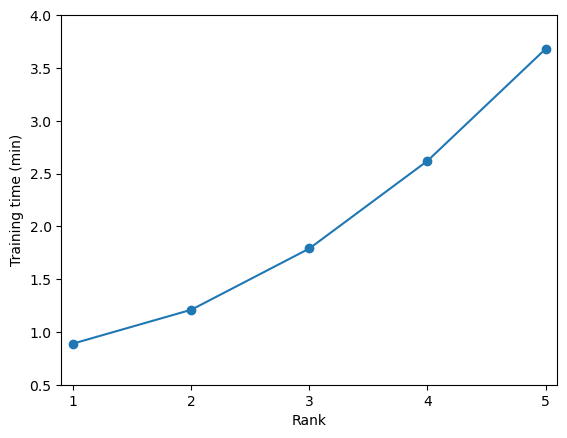}}
	\subfigure[\# of parameters to express tensor $\gT$ vs rank $d$. \label{fig:param}]{\includegraphics[width=0.48\columnwidth]{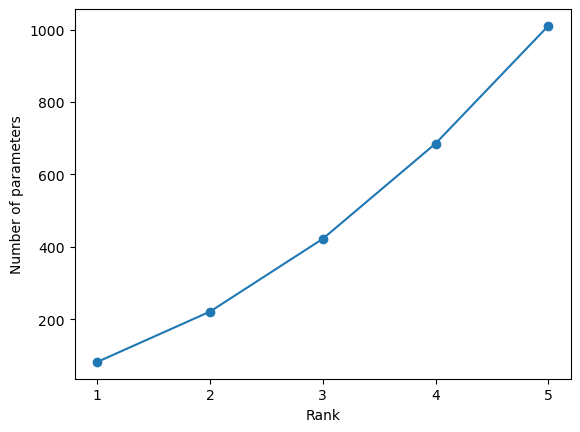}}
	\vspace{-10px}
	\caption{Effect of the hyper-parameter $d$ (rank).}
	\vspace{-10px}
\end{figure*}

We compare TENSUS with 
the same baselines in Section~\ref{sec:node}, 
and GEMS~\cite{han2020gems}, 
which adopts a parallel genetic algorithm to search meta-structures for recommendation task and propose an attention based method to fuse information from meta-structures. 
We follow 
the data preprocessing and experimental settings 
in 
DiffMG~\cite{ding2021diffmg}. 
The embedding dimension is 64.
We use the Adam optimizer~\cite{adam} 
with its hyper-parameters (learning rate, weight decay and input dropout) 
tuned according to the validation set.
For performance evaluation, we use 
the average AUCs
over 10 runs with different random seeds. 

Table~\ref{tab:rec} shows the AUCs.
Similar to the node classification task, HAN and MAGNN perform worse 
than those methods that can learn meta-paths or meta-graphs, 
i.e. GEMS, GTN, HGT, and DiffMG, 
which demonstrates the necessity of learning meta-graphs. 
The proposed parameterization TENSUS can improve upon DiffMG 
by the topology-aware tensor decomposition. 

We also plot the learning curves of different methods in Fig.~\ref{fig:db} 
for the Douban dataset,
which demonstrates that our parameterization TENSUS can help find 
meta-paths or meta-graphs that have a more stable performance.

\subsection{Ablation Study}

\subsubsection{Visualization of learned meta-graphs} 

We first visualize the learned meta-graphs from DiffMG and TENSUS in Figures~\ref{fig:dmgs} 
and \ref{fig:tens}, respectively. 
The ``I'' operation in these figures indicate identity operation, i.e. the node embeddings are directly added 
to later results. 
From these two figures, we can see that TENSUS identifies meta-graphs that are less 
complex and achieve a better performance, 
as is demonstrated in Table~\ref{tab:rec}. 

\subsubsection{Impact of topology information}
\label{sec:exp:topo}
\begin{figure}[ht]
	\centering
	\vspace{-10px}
	\subfigure[Validation AUC vs \# of parameters. \label{fig:cpdv}]{\includegraphics[width=0.48\columnwidth]{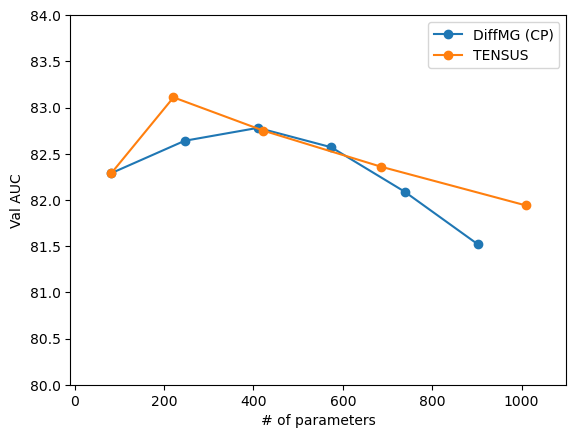}}
	\subfigure[Test AUC vs \# of parameters. \label{fig:cpdt}]{\includegraphics[width=0.48\columnwidth]{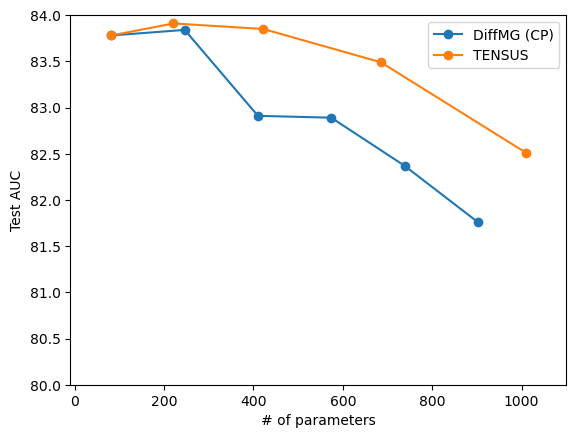}}
	\vspace{-10px}
	\caption{Comparison between CP and TENSUS.}
	\vspace{-10px}
	\label{fig:compcp}
\end{figure}

In this section, we compare the performance of our topology-aware decomposition
with the CP decomposition under similar number of parameters 
by controlling the rank of these two decompositions. 
Results are shown in Fig.~\ref{fig:compcp}. 
From the figure, we can see that our proposed method 
achieves a better result than CP with similar number of parameters. 
This demonstrates that the performance gain of our proposed method
comes from modeling topological structures rather than more parameters.

\subsubsection{Effect of Hyper-Parameter $d$} 
\label{ssec:hyp}
Finally, we study the effect of the 
rank $d$, which
controls the expressive power of the proposed parameterization. 
Figures~\ref{fig:rank} and \ref{fig:rankt}
show the validation and test AUC of TENSUS 
with different $d$ on the Douban dataset.  
For easy reference, we also show the performance of
DiffMG~\cite{ding2021diffmg}. As can be seen,
TENSUS outperforms DiffMG 
over a range of $d$'s.
As expected,
a $d$ too small 
limits the expressive power, 
while a $d$ too large can make optimization difficult.
A large $d$ also increases the training time as well as the number of parameters, 
as is shown in Figures~\ref{fig:time} and \ref{fig:param}, respectively. 
However, the total number of parameters here is only about 1000 even when $d$ is set to 5, 
which is very small if compared with the parameters in a GNN model.

\section{Conclusion}

In this paper, we study the problem of learning suitable meta-graphs 
on a given heterogeneous graph.  
Instead of modeling the edges in a meta-graph independently, 
which corresponds to CP decomposition, 
we propose to impose how node embeddings are computed
onto the tensor of weights for meta-graphs.
The decomposition is inspired by the parsimonious tensor train decomposition, 
and can be trained efficiently together with another graph data mining model. 
Empirical results on three diverse tasks demonstrate that the proposed method can find better meta-graphs and outperforms the state-of-the-arts. 

\ifCLASSOPTIONcompsoc
  \section*{Acknowledgments}
  Q. Yao is supported by NSF of China (No. 92270106).
\else
  \section*{Acknowledgment}
  Q. Yao is supported by NSF of China (No. 92270106).
\fi


\ifCLASSOPTIONcaptionsoff
  \newpage
\fi

\bibliographystyle{IEEEtran}
\bibliography{bib}

\begin{IEEEbiography}{Hansi Yang}
Biography text here.
\end{IEEEbiography}

\begin{IEEEbiographynophoto}{Peiyu Zhang}
	Biography text here.
\end{IEEEbiographynophoto}

\begin{IEEEbiographynophoto}{Quanming Yao}
Biography text here.
\end{IEEEbiographynophoto}

\newpage

\appendices
\section{Proof for Proposition~\ref{pr:norm}}
\label{app:fail}

\begin{proof}
By the definition of softmax operation, we should have 
\begin{align*}
\gW_{0,k}(i_{0,k}, r_k) & > 0, \sum_{i_{0,k}=1}^C \gW_{0,k}(i_{0,k}, r_k) = 1, \\
\gW_{j, k}(i_{j,k}, r_j, r_{k}) & > 0, \sum_{i_{j,k}=1}^C \gW_{j, k}(i_{j,k}, r_j, r_{k}) = 1,  \\
\gW_{0, N+1}(i_{0,N+1}) & > 0, \sum_{i_{0,N+1}=1}^C \gW_{0, N+1}(i_{0,N+1}) = 1,  \\
\gW_{j, N+1}(i_{j, N+1}, r_j) & > 0, \sum_{i_{j,N+1}=1}^C \gW_{j, N+1}(i_{j, N+1}, r_j) = 1,
\end{align*}
hold for all $r_j, r_k = 1, \dots, d$. Then obviously we have $\mathcal{T}(i_{0,1}, \dots, i_{N, N+1}) > 0$. 
And to prove $\sum_{i_{0,1}=1}^C \dots \sum_{i_{N, N+1}=1}^C \mathcal{T}(i_{0,1}, \dots, i_{N, N+1}) = 1$, 
we first re-write the summation to follows: 
\begin{align*}
& \sum_{i_{0,1}=1}^C \dots \sum_{i_{N, N+1}=1}^C \mathcal{T}(i_{0,1}, \dots, i_{N, N+1}) \\
& = \sum_{i_{0,1}=1}^C \dots \sum_{i_{N, N+1}=1}^C  \frac{1}{d^N} \sum\nolimits_{r_{1}=1}^{d}\cdots\sum\nolimits_{r_{N}=1}^{d} \notag \\
& \gW_{0,1}(i_{0,1}, r_{1}) \cdots \gW_{0,N}(i_{0,N}, r_{N}) \notag \\
& \gW_{1,2}(i_{1,2}, r_1, r_2) \cdots \gW_{N-1, N}(i_{N-1, N}, r_{N-1}, r_N) \notag \\
& \gW_{0,N+1}(i_{0,N+1}) \gW_{1,N+1}(i_{1,N+1}, r_{1})  \cdots \gW_{N,N+1}(i_{N,N+1}, r_{N}) \\
& = \frac{1}{d^N} \sum\nolimits_{r_{1}=1}^{d}\cdots\sum\nolimits_{r_{N}=1}^{d} \notag \\
& \left( \sum_{i_{0,1}=1}^C \gW_{0,1}(i_{0,1}, r_{1}) \right) \cdots \left( \sum_{i_{0,N}=1}^C \gW_{0,N}(i_{0,N}, r_{N}) \right) \notag \\
& \left( \sum_{i_{1,2}=1}^C  \gW_{1,2}(i_{1,2}, r_1, r_2)\right)  \cdots \left( \sum_{i_{N-1,N}=1}^C \gW_{N-1, N}(i_{N-1, N}, r_{N-1}, r_N) \right) \notag \\
& \left( \sum_{i_{0,N+1}=1}^C \gW_{0,N+1}(i_{0,N+1}) \right) \left( \sum_{i_{1,N+1}=1}^C \gW_{1,N+1}(i_{1,N+1}, r_{1}) \right)  \\
& \cdots \left( \sum_{i_{N,N+1}=1}^C \gW_{N,N+1}(i_{N,N+1}, r_{N}) \right) \\
\end{align*}
Then from the properties of softmax operation, we have:
\begin{align*}
& \sum_{i_{0,1}=1}^C \dots \sum_{i_{N, N+1}=1}^C \mathcal{T}(i_{0,1}, \dots, i_{N, N+1}) \\
& = \frac{1}{d^N} \sum\nolimits_{r_{1}=1}^{d}\cdots\sum\nolimits_{r_{N}=1}^{d} 1 \cdots 1 \\
& = \frac{1}{d^N} \cdot d^N =1 \\
\end{align*}
where the $d^{N}$ comes from the $N$ summations of $r_1, \dots, r_N$ from $1$ to $d$. And this concludes our proof. 
\end{proof}

\end{document}